\documentclass[letterpaper, 10 pt, journal, twoside]{IEEEtran}
\ifCLASSINFOpdf
\else
\fi
\usepackage{cite}
\usepackage{amsmath,amssymb,amsfonts}
\usepackage{graphicx}
\usepackage{textcomp}
\usepackage{xcolor}
\usepackage{gensymb}
\usepackage{multirow}
\usepackage{subcaption}
\usepackage{commath} 
\usepackage{ushort} 
\usepackage{url} 

\usepackage{epsfig} 
\usepackage{balance} 

\usepackage{adjustbox}
\usepackage{booktabs, makecell, tabularx}
\usepackage{siunitx}
\setcellgapes{3.5pt}
\usepackage{threeparttable}

\usepackage{algorithm} 
\usepackage[noend]{algpseudocode} 
\usepackage{footnote} 
\usepackage{float} 
\usepackage{kotex}
\usepackage{wrapfig}

\makeatletter
\let\NAT@parse\undefined
\makeatother
\usepackage[breaklinks,hidelinks]{hyperref}

\begin{document}
%
\title{
TempFuser: Learning Agile, Tactical, and Acrobatic Flight Maneuvers Using a Long Short-Term Temporal Fusion Transformer
}
%
%
%


\author{Hyunki Seong and David Hyunchul Shim

\thanks{Manuscript received: April 27, 2024; Revised July 31, 2024; Accepted September 9, 2024. This paper was recommended for publication by Editor Aleksandra Faust upon evaluation of the Associate Editor and Reviewers' comments.
This research was supported by a research program (No. UE191120RD) through the Agency for Defence Development.\\ \textit{(Corresponding author: David Hyunchul Shim)}} 
\thanks{Hyunki Seong and David Hyunchul Shim are with the School of Electrical Engineering, Korea Advanced Institute of Science and Technology, Daejeon, South Korea.
        {\tt\small \{hynkis,hcshim\}@kaist.ac.kr}
        }%
\thanks{Digital Object Identifier (DOI): see top of this page.}
}

%
%

\markboth{IEEE Robotics and Automation Letters. Preprint Version. Accepted SEPTEMBER, 2024}
{Seong \MakeLowercase{\textit{et al.}}: TempFuser: Long Short-Term Temporal Fusion Transformer} 

%



\maketitle

\begin{abstract}
Dogfighting is a challenging scenario in aerial applications that requires a comprehensive understanding of both strategic maneuvers and the aerodynamics of agile aircraft. The aerial agent needs to not only understand tactically evolving maneuvers of fighter jets from a long-term perspective but also react to rapidly changing aerodynamics of aircraft from a short-term viewpoint. In this paper, we introduce TempFuser, a novel long short-term temporal fusion transformer architecture that can learn agile, tactical, and acrobatic flight maneuvers in complex dogfight problems. Our approach integrates two distinct temporal transition embeddings into a transformer-based network to comprehensively capture both the long-term tactics and short-term agility of aerial agents. By incorporating these perspectives, our policy network generates end-to-end flight commands that secure dominant positions over the long term and effectively outmaneuver agile opponents. After training in a high-fidelity flight simulator, our model successfully learns to execute strategic maneuvers, outperforming baseline policy models against various types of opponent aircraft. Notably, our model exhibits human-like acrobatic maneuvers even when facing adversaries with superior specifications, all without relying on prior knowledge. Moreover, it demonstrates robust pursuit performance in challenging supersonic and low-altitude situations. Demo videos are available at \url{https://sites.google.com/view/tempfuser}.
\end{abstract}

\begin{IEEEkeywords}
Aerial Systems: Applications, Reinforcement Learning, Machine Learning for Robot Control
\end{IEEEkeywords}

%
\IEEEpeerreviewmaketitle

\section{Introduction}
\label{sec:introduction}
\IEEEPARstart{A}{ir-to-air} combat is the tactical art of maneuvering an agile flight agent to reach a position to aim at an opponent. It is also known as dogfighting, as in most cases, each fighter jet pursues the tail of the other in short-range combat situations.
Dogfights present three key challenges in aerial applications: 
1) They are highly interactive and constantly evolving situations where each agent attempts to maximize its positional advantage. This dynamic nature requires continuous assessment and rapid response to changes.
2) These scenarios demand tactics and accuracy, especially when trying to position a high-speed opponent within a tight effective damage range. Precision in execution is critical to success in such high-stakes environments.
3) Dogfights take place in a 3D environment governed by complex aerodynamics and are subject to safety altitude restrictions to avoid ground collisions. Navigating this space effectively adds another layer of complexity to the already demanding conditions.

For successful dogfights, the agent requires a combination of situational awareness, strategic planning, and maneuverability from long and short-term perspectives.
Firstly, the agent has to plan its tactical position by understanding the opponent's long-term trajectories. Naive chasing after the adversary's immediate positions may provide a temporary advantage, but it can eventually leave itself in a vulnerable position later.
Therefore, the agent should constantly assess the opponent's long-term maneuvers, react to their actions, and strategically position itself to gain an advantage over the adversary.
Secondly, the agent needs to have the ability to comprehend the agile maneuverability of the aircraft from a short-term dynamics perspective. Modern fighter jets are engineered to possess high maneuverability, enabling them to swiftly alter direction and speed, resulting in aerobatic movements in the engagement situation. Thus, to maintain an advantageous position over the opponent, the agent should promptly grasp both the opponent's agile movements and the agent's own potential maneuvers from a dynamic perspective.

\begin{figure}[t]
\centering
\includegraphics[width=0.45\textwidth]{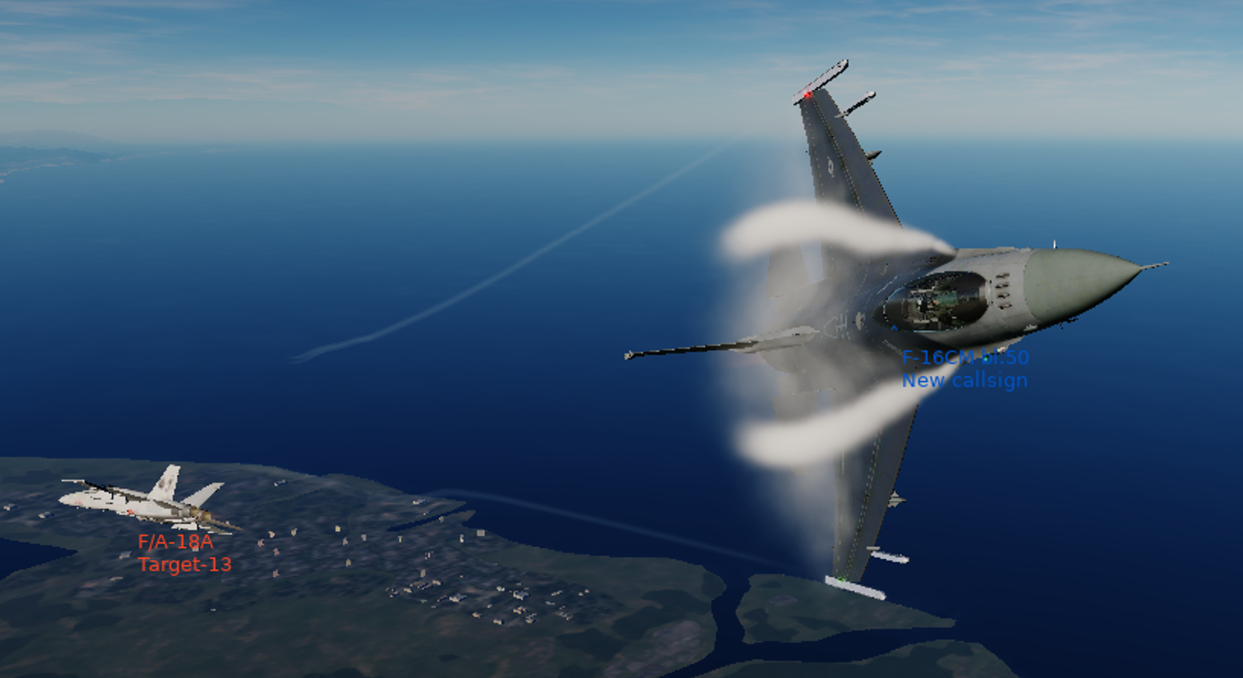}
\caption{
Dogfighting requires a combination of agility, tactics, and aerobatics to secure dominant positions over the opponent in complex airborne scenarios.
}
\label{fig:intro}
\vspace{-1.5em}
\end{figure}
In order to handle those challenges, conventional approaches design rule-based policies that employ appropriate Basic Fighter Maneuvers (BFMs) derived from human pilot experience\cite{burgin1988rule,shin2018autonomous,yang2024manual}. However, they require complex handcrafted rule sets and rely heavily on heuristics for choosing proper flight maneuvers. This leads to a lack of generality for various aircraft and flight policies.
On the other hand, modern deep learning-based approaches, especially those based on deep reinforcement learning (DRL), implement complex policies through data-driven and experience-based schemes\cite{yang2019maneuver,fan2022air,hu2022autonomous,pope2021hierarchical,kong2020maneuver,yoo2022deep,bae2023deep}. Without supervision of demonstrations, DRL allows for the optimization of policies through environment interactions using reward functions.
However, despite the success of numerous traditional networks in aerial applications, developing a policy network that can effectively understand and execute strategic maneuvers of high-speed fighter jets still remains a challenge.

\begin{figure}[t]
\centering
\includegraphics[width=0.46\textwidth]{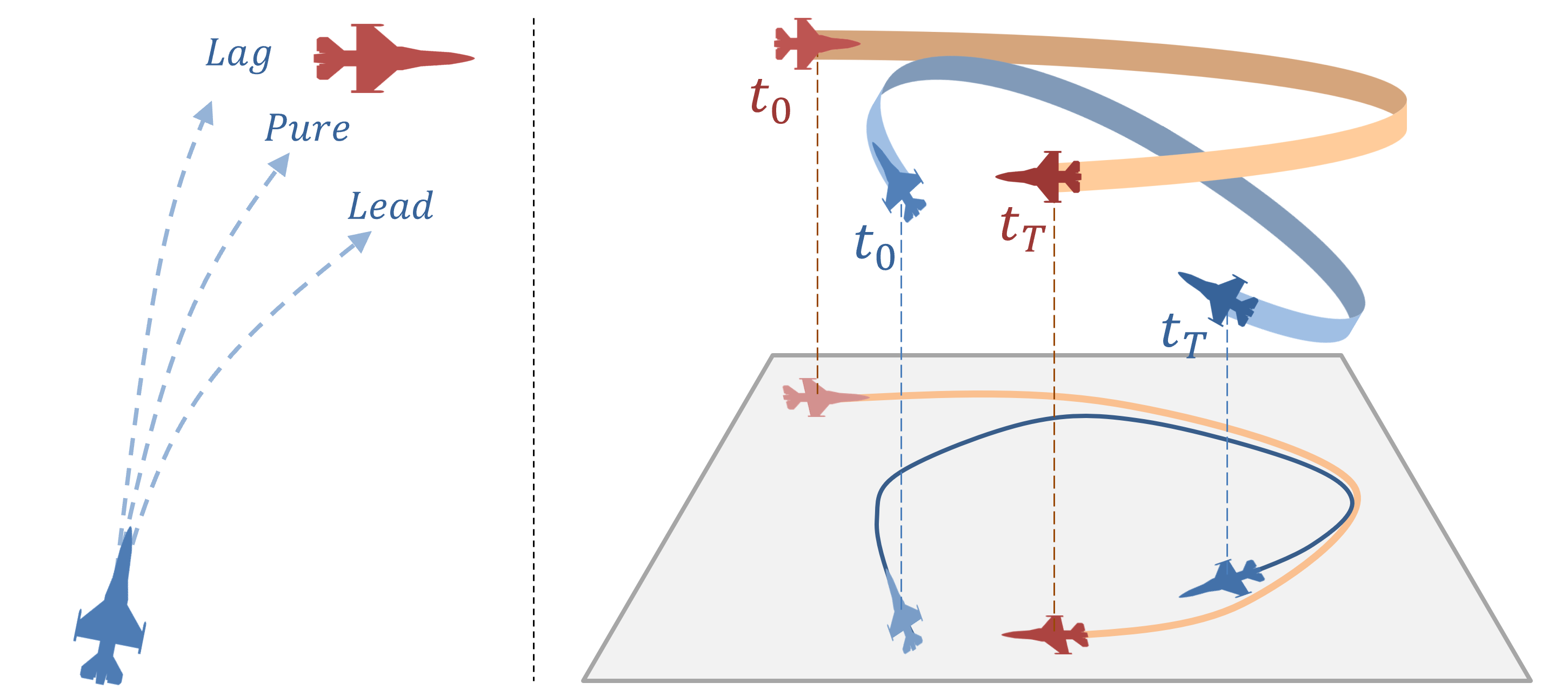}
\caption{
Human pilot's tactical maneuvers: pursuit strategies (left) and an out-of-plane maneuver, \textit{`Low Yo-Yo'} (right).
}
\label{fig:tactical}
\vspace{-1.5em}
\end{figure}

In this paper, we propose \textit{TempFuser}, a transformer-based\cite{dosovitskiy2020image} network that integrates long and short-term temporal transition embeddings to learn agile, tactical, and acrobatic maneuvers in airborne scenarios. The network employs two separate 
Long Short-Term Memory (LSTM) units \cite{hochreiter1997long} to extract features of the overall flight maneuvers and instantaneous dynamical transitions, respectively, from the corresponding temporal trajectories. Additionally, a transformer encoder extracts global contexts from the temporal features, which reflect the opponent's tactical and dynamical characteristics. 
By integrating these distinct perspectives into the policy inference, TempFuser gains a comprehensive understanding of both the tactical situation and agile flight dynamics. This enables it to execute strategic end-to-end flight controls to outperform the opponent in dogfight scenarios.

We tackle the aerial dogfight problem in the Digital Combat Simulator (DCS)\cite{DCS}, considered one of the most realistic environments. It is widely used for pilot training due to its high-fidelity aircraft dynamics, comprehensive environmental factors, and wide range of high-quality airborne scenarios\cite{DCSTerms}.
We formulate the dogfighting problem as a DRL framework and design an energy-embedded reward function that enables the agent to discover and learn acrobatic flight maneuvers without prior supervision.
We extensively train and validate our network with a variety of opponent aircraft, including the F-15E, F-16, F/A-18A, and Su-27\cite{bongers2014measuring}, in our complex 3D airborne environments.
As a result, our TempFuser demonstrates challenging agile flight maneuvers in an end-to-end manner and outperforms various opponent aircraft, including those with superior specifications. Additionally, it exhibits robust pursuit performance at low altitudes and high-speed flight scenarios above Mach 1.

The key contributions of our research are as follows:
\begin{itemize}
    \item 
    We design TempFuser, a novel long short-term temporal fusion transformer to learn agile and tactical flight.
    
    \item 
    We formulate a DRL framework that includes an energy-embedded reward function tailored for aerial dogfighting in a high-fidelity airborne environment.
    
    \item
    We extensively evaluate our model against diverse types of opponents, demonstrating acrobatic maneuvers, outperforming superior-spec aircraft, and achieving robust pursuit in supersonic conditions, all without explicit prior knowledge.
\end{itemize}

\begin{figure}[t]
\centering
\includegraphics[width=0.46\textwidth]{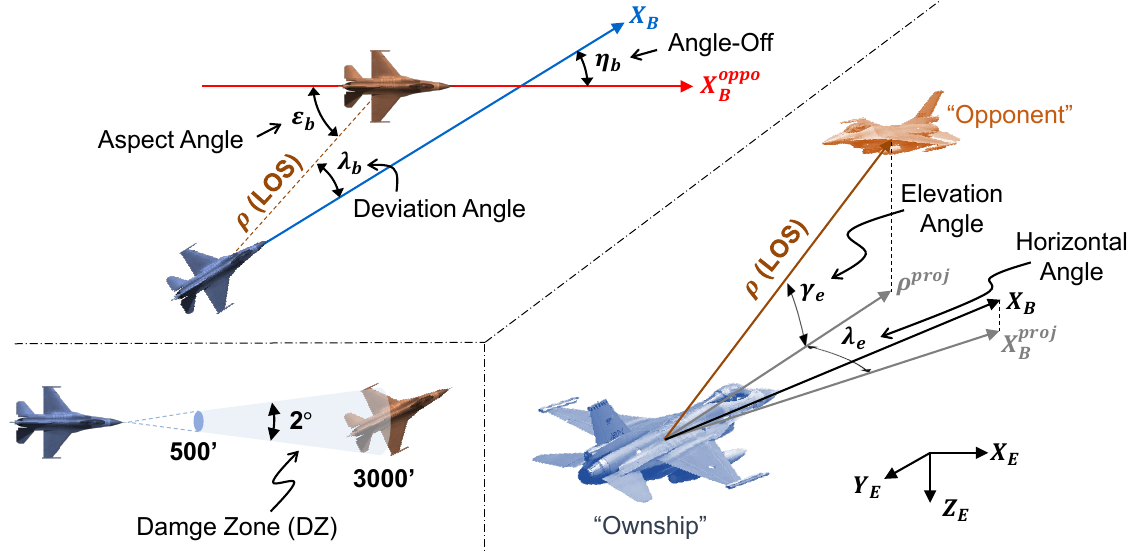}
\caption{
Geometries for aerial dogfights.
}
\label{fig:geometry}
\vspace{-1.5em}
\end{figure}
\section{Related Works}
\label{sec:relatedworks}

\subsection{Conventional Approaches}
Based on human pilot tactics (Fig. \ref{fig:tactical}) \cite{shaw1985fighter}, most of the conventional studies used rule-based heuristics as a design approach.
They proposed guidance laws with BFMs for the selection of pursuit strategies \cite{burgin1988rule,you2015design} and offensive/defensive maneuvers \cite{shin2017design,shin2018autonomous,yang2024manual} in air-to-air scenarios. Although heuristic-based methods were efficient in practice, they often suffered from the need for manual adjustment of parameters and flexibility issues in complex aerial environments.
Several theoretical approaches utilized optimization methods, such as approximate dynamic programming (ADP) or differential game theory. The ADP provides a fast response by efficiently approximating the optimal policy\cite{mcgrew2010air, wang2020influence}. The differential game methods designed a scoring function matrix to generate optimal maneuvers\cite{ardema1987approach, park2016differential}. However, they often require a finite action representation for real-time computation, which is unsuitable for maneuvering in a large action space.

\subsection{Learning-based Approaches}
Recent Deep RL-based studies can be categorized into two parts: hierarchical and end-to-end approaches.
Hierarchical approaches involve a hierarchical structure with a high-level policy and a low-level controller. The policy infers discretized high-level actions in terms of maneuvers or tactics, while the controller computes low-level commands based on the actions.
They construct a maneuver library\cite{austin1990game} expanded by a set of basic control values to choose a flight maneuver\cite{zhang2018research,yang2019maneuver,hu2021application,fan2022air,hu2022autonomous}. Alternatively, they configure multiple sub-policies to select a proper strategy based on the current context of the combat geometry\cite{yang2020evasive,piao2020bvr, pope2021hierarchical}.
Despite their efficient policy optimization within a small search space, they are limited to pre-defined maneuvers and handcrafted strategies, resulting in a lack of generality.
On the other hand, end-to-end methods directly map the geometry-based states into flight control actions.
They design and train neural networks based on MLP\cite{yang2019uav,kong2020maneuver, yoo2022deep} or LSTM\cite{bae2023deep}, which enable policy learning from experiential data and reward functions, eliminating the need for hand-designed components.
However, it is less explored in the literature to develop an end-to-end policy network that can comprehend the tactical features of agile fighter jets without explicit prior knowledge.

\section{Flight Geometries}
\label{sec:problem_statement}
The geometrical relationships between the aircraft in dogfights are illustrated in Fig. \ref{fig:geometry}. For simplicity, the ego and opponent aircraft are referred to as the ownship and the opponent, respectively.
The line-of-sight (LOS) vector $\boldsymbol{\rho}$ 
 denotes a vector from the ownship to the opponent.
The deviation and aspect angle ($\lambda_b, \epsilon_b$) represent the angle between $\boldsymbol{\rho}$ and each heading vector of the ownship ($\boldsymbol{X_b}$) and opponent ($\boldsymbol{X^{oppo}_{b}}$), respectively.
The angle-off $\eta_b$ indicates the angular difference between the heading vector of the ownship and opponent.
We further define the horizontal and elevation angles ($\lambda_E, \gamma_E$) that represent the horizontal and vertical deviation angles that can be obtained from Eq. \ref{eq:geometry_angle}: 
\begin{equation}
    \label{eq:geometry_angle}
    \gamma_e = \text{cos}^{-1} \left[\frac{\boldsymbol{\rho} \cdot \boldsymbol{\rho^{proj}}}{|\boldsymbol{\rho}| |\boldsymbol{\rho^{proj}}|}\right], 
    \lambda_e = \text{cos}^{-1} \left[\frac{\boldsymbol{X^{proj}_B} \cdot \boldsymbol{\rho^{proj}}}{|\boldsymbol{X^{proj}_B}| |\boldsymbol{\rho^{proj}}|}\right],
\end{equation}
where the superscript $\boldsymbol{proj}$ indicates a projected vector with respect to the global plane $\boldsymbol{XY_E}$.
In air combat research, the geometric area where the agent can effectively inflict damage can be defined as a damage zone (DZ), which is a two-degree spherical cone truncated at a distance range of \SI[group-separator={,},group-minimum-digits=4]{500} to \SI[group-separator={,},group-minimum-digits=4]{3000}{ft} from the ego aircraft\cite{pope2021hierarchical}.
\section{Methodologies}
\label{sec:methodologies}
\subsection{State and Action Representation}
The state space is represented by three elements: $x^{own}_{t}$, $x^{oppo}_{t}$ and $a_{t-1}$.
The first element $x^{own}_{t}\!\in\!\mathbb{R}^{13}$ has aircraft flight states of the ownship as follows: 
\begin{equation}
    \label{eq:state_own}
    x^{own}_{t} = \{v^{b}_{x}, a^{b}_{x}, a^{b}_{z}, \phi_{b}, \theta_{b}, \Dot{\phi}_{b}, \Dot{\theta}_{b}, \Dot{\psi}_{b}, \alpha_{b}, \beta_{b}, E, z_{e}, v^{e}_{z}\}
\end{equation}
where $v^{b}_{x}$ is the true airspeed which is aligned with the axis of the relative wind in aircraft.
$a^{b}_{x}, a^{b}_{z}$ are the longitudinal and vertical acceleration, and $\phi_{b}, \theta_{b}$ are the roll and pitch angle in the body frame.
$\Dot{\phi}_{b}, \Dot{\theta}_{b}, \Dot{\psi}_{b}$ denote the rate of the roll, pitch, and yaw angle.
$\alpha_b$ and $\beta_b$ indicate the angle of attack (AoA) and side slip angle.
$E \! = z_e \! + (v^{b}_{x})^2/(2g)$ represents the specific energy of the ownship, where $z_e$ is the altitude and $g$ is the gravitational acceleration.
$v^{e}_{z}$ is the vertical velocity in the earth frame.
Another state element $x^{oppo}_{t} \in \mathbb{R}^{16}$ has geometric and dogfight-related state variables as 
\begin{equation}
    \label{eq:state_oppo}
    x^{oppo}_{t} = \{P^{bo}_{e}, P^{bo}_{b}, O^{bo}_{b}, \lambda_{b}, \epsilon_{b}, \eta_{b}, \lambda_{e}, \gamma_{e}, \xi_{own}, \xi_{oppo}\},
\end{equation}
where $P^{bo}_e, P^{bo}_b \! \in \! \mathbb{R}^3$ denote the opponent's relative position in the earth and body frame, respectively, $O^{bo}_b \! \in \! \mathbb{R}^3$ indicates the opponent's relative orientation to the body frame. $\xi_{own}, \xi_{oppo}$ refer to the life scores of the ownship and opponent.
All the state elements are normalized to $[-1,1]$.
We also append the previous action $a_{t-1} \! \in \! \mathbb{R}^4$ to the state, allowing our policy to infer the ownship's underlying dynamics using aircraft states and past action\cite{peng2018sim}.
The total features ${x}_{t} \! = \{x^{own}_{t}, x^{oppo}_{t}, a_{t-1}\}$ result in a 33D state space.

The action space is represented by four control commands:
\begin{equation}
    \label{eq:action}
    {a}_{t} = \{u_{\phi}, u_{\theta}, u_{\psi}, u_{\tau}\}
\end{equation}
where $u_{\phi}, u_{\theta}, u_{\psi}, u_{\tau}$ specify the aileron, elevator, rudder, and thrust commands of the ownship in continuous space.

\subsection{Long Short-Term Temporal Fusion Transformer}
\begin{figure}[t]
\centering
\includegraphics[width=0.48\textwidth]{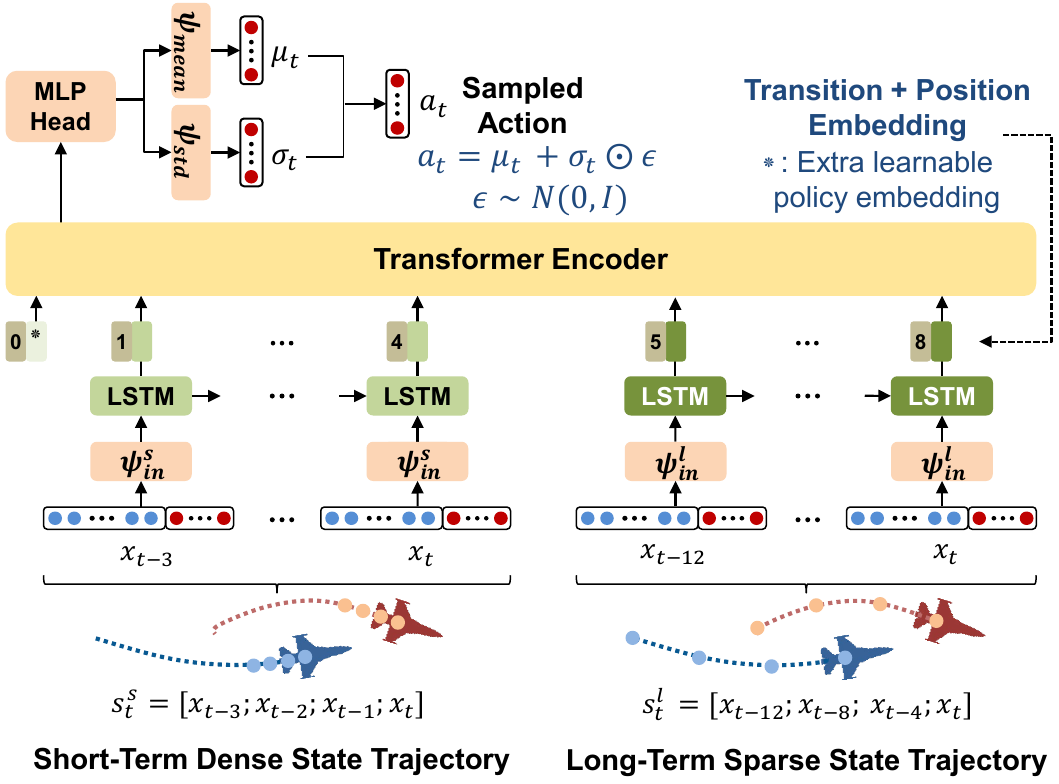}
\caption{
Overview of the TempFuser architecture for the policy network with the following example sets $n_s = n_l = \Delta t = 4$.
}
\label{fig:tempfuser}
\vspace{-1em}
\end{figure}

A schematic diagram of TempFuser is shown in Fig. \ref{fig:tempfuser}.
We configure long short-term state trajectories to represent tactical and dynamical state transitions.
The short-term dense trajectory $s^s_t \! \in \mathbb{R}^{n_s \! \times 33}$ is a state history of length $n_s$ with a single-step interval that includes the current state $x_t$.
Since $s^s_t$ has the same time resolution as the environment, the dense trajectory represents the dynamic state transition information (Eq. \ref{eq:state_short}).
On the other hand, the long-term sparse trajectory $s^l_t \! \in \mathbb{R}^{n_l \! \times 33}$ is a state history of length $n_l$ with multi-step interval $\Delta t$. It exhibits different state transitions from aircraft dynamics, containing the overall maneuver-level information (Eq. \ref{eq:state_long}).
In this work, we set $n_s\!=\!n_l\!=\!\Delta t\!=\!8$ to observe long-term trajectories with sufficient maneuver-level features without extensive parameter search.
\begin{flalign}
    \label{eq:state_short}
    {s}^{s}_{t} &= [x_{t-(n_s\!-1)}; x_{t-(n_s\!-2)}; \cdots ; x_{t-1}; x_t] \\
    \label{eq:state_long}
    {s}^{l}_{t} &= [x_{t-(n_l\!-1)\Delta t}; x_{t-(n_l\!-2)\Delta t}; \cdots ; x_{t-\Delta t}; x_t]
\end{flalign}

To handle the distinct temporal representations, we employ two LSTM-based input embedding pipelines.
We first encode each temporal representation through a linear layer $\psi^i_{in} \in \mathbb{R}^{33 \times d}, i = \{l,s\}$ with ReLU nonlinearity. We then sequentially process each encoded trajectory feature with its corresponding LSTM. This generates hidden outputs for each trajectory, which we configure as a temporal transition embedding (Eq. \ref{eq:lstm}).
By employing two individual LSTMs, we incorporate distinct sequential relational inductive biases\cite{battaglia2018relational} associated with both dense and sparse state transitions into the input embeddings. As a result, the agent extracts not only the instantaneous physical properties but also the comprehensive features of the maneuvers from observations.

As all layers use the same layer size $d$, the two pipelines encode the input trajectories to long and short-term transition embeddings, $h_l \!\in\! \mathbb{R}^{n_l \!\times d}$ and $h_s \!\in\! \mathbb{R}^{n_s \!\times d}$, respectively.
\begin{flalign}
    \label{eq:lstm}
    h_i = \text{LSTM} (\text{ReLU}(\psi^i_{in}(s^i_t))), \quad i = \{l,s\}
\end{flalign}

We leverage a transformer encoder \cite{dosovitskiy2020image} that can learn global context to fuse two distinct transition embeddings. Before feeding the embeddings into the encoder, we concatenate them into a single sequential embedding and prepend a learnable policy token $z_{policy} \in \mathbb{R}^{1 \times d}$. This token serves as a 1-D representation vector to summarize the sequence input and represent information about the multi-temporal state transition.
Additionally, we add another learnable position embedding $z_{pos} \! \in \! \mathbb{R}^{(n_s+n_l+1) \times d}$ to the sequential embedding to provide a positional feature for each element (Eq. \ref{eq:embedding}).
\begin{flalign}
    \label{eq:embedding}
    z_0 &= [z_{policy}; h^{1}_s; h^{2}_s; ...; h^{n_s}_s; h^{1}_l; h^{2}_l; ...; h^{n_l}_l] + z_{pos}
\end{flalign}

Following \cite{dosovitskiy2020image}, the transformer encoder consists of a multi-head self-attention (MSA) block and an MLP block, as well as layer normalizations (LN) and residual connections (Eq. \ref{eq:transformer-msa}, \ref{eq:transformer-mlp}). The MLP has two FC layers with GeLU activation. The encoders are stacked $L$ times, increasing the capacity of the transformer network. The first element in the output of the last encoder ($z^{0}_L$), corresponding to the policy token, is derived as the final output $y \in \mathbb{R}^{d}$ through another MLP head composed of a linear layer and LN (Eq. \ref{eq:transformer-out}).
\begin{flalign}
    \label{eq:transformer-msa}
    z'_k &= \text{MSA} (\text{LN} (z_{k-1})) + z_{k-1}, \quad k = 1, ..., L \\
    \label{eq:transformer-mlp}
    z_k &= \text{MLP} (\text{LN} (z'_k)) + z'_k, \quad\quad\quad  k = 1, ..., L \\
    \label{eq:transformer-out}
    y &= \text{MLP} (\text{LN} (z^{0}_L))
\end{flalign}

TempFuser computes the action $a_t \in \mathbb{R}^{4}$ in continuous space using a squashed Gaussian policy\cite{haarnoja2018softapplication}. The mean and standard deviation of the action $\mu_t, \sigma_t$ are computed through linear projections $\psi_{mean}, \psi_{std} \in \mathbb{R}^{d \times 4}$, respectively (Eq. \ref{eq:gaussian-policy}). During training, a stochastic action is sampled from these two values using the reparameterization trick of the Gaussian policy \cite{schulman2015gradient}. During evaluation, only the mean is used to derive a deterministic action. We use a nonlinear squashing function (tanh) for the action to be bounded within a finite range of [-1, 1] (Eq. \ref{eq:squashed_gaussian}).
\begin{flalign}
    \label{eq:gaussian-policy}
    \mu_t &= \psi_{mean}(\text{LN} (y)), \quad \sigma_t = \psi_{std}(\text{LN} (y)) \\
    \label{eq:squashed_gaussian}
    {a}_t &= \tanh{(\mu_{t} + \sigma_{t} \odot \xi)}, \quad \xi \sim N(0, I).
\end{flalign}

\subsection{Soft Actor-Critic with State Trajectories}
We construct an actor-critic architecture to optimize our policy network, following the Soft Actor-Critic (SAC) method which incorporates the clipped double-Q trick\cite{haarnoja2018softapplication}.
In this setup, the TempFuser-based policy $\pi_{\phi}$, parameterized by $\phi$, serves as the actor. Additionally, we configure two Q-function networks $Q_{\theta_1}$ and $Q_{\theta_2}$, parameterized by $\theta_1$ and $\theta_2$, respectively, as the critics.
The Q-network also incorporates the TempFuser-based pipeline, that includes two distinct LSTM-based input embeddings and a transformer encoder. To compute the Q-value, a state-action value, we concatenate the action $a_t$ with the transformer output. The Q-value is then derived using three additional linear projections. All linear layers in the critic network employ ReLU nonlinearity.

\begin{algorithm}[t]
\caption{SAC Training Algorithm with State Trajectories}
\label{algo:sac}
\small
\textbf{Input: } $\phi, \theta_{1}, \theta_{2} $ \\
\textbf{Output: } \text{Optimized parameters } $\phi, \theta_{1}, \theta_{2} $

\begin{algorithmic}[1]
\State $ \Bar{\theta}_{1} \leftarrow \theta_{1}, \Bar{\theta}_{2} \leftarrow \theta_{2}, D \leftarrow \emptyset, \mathcal{T}^{\dagger} \leftarrow \emptyset$

\For{each episode}
    \While{$n_l \Delta t$ steps}
        \State Execute action ${a}_{init}$ and observe $x^{own}_{t+1}, x^{oppo}_{t+1}, r_t$
        \State $\mathcal{T}.\text{enqueue}(\{x^{own}_{t+1}, x^{oppo}_{t+1}, a_{init}\})$
    \EndWhile
    \State Initialize ${s}^{s}_{t}, {s}^{l}_{t}$ by indexing from $\mathcal{T}$
    \While{not done}
        \State $ {a}_{t} \sim \pi_{\phi}({a}_{t} | {s}^{s}_{t}, {s}^{l}_{t})$
        \State Execute action ${a}_{t}$ and observe $x^{own}_{t+1}, x^{oppo}_{t+1}, r_t$
        \State $ \mathcal{T}.\text{dequeue}()$ and $\mathcal{T}.\text{enqueue}(\{x^{own}_{t+1}, x^{oppo}_{t+1}, a_{t}\})$
        \State $ {s}^{s}_{t+1} \leftarrow \{\mathcal{T}[n_l \Delta t\! -(n_s\! -1)]; \mathcal{T}[n_l \Delta t\! -(n_s\! -2)]; \cdots; \newline
        \hspace*{6em} \mathcal{T}[n_l \Delta t -1]; \mathcal{T}[n_l \Delta t]\} $
        \State $ {s}^{l}_{t+1}\! \leftarrow \{\mathcal{T}[\Delta t]; \mathcal{T}[2 \Delta t]; \cdots ; \mathcal{T}[(n_l\! -1) \Delta t]; \mathcal{T}[n_l \Delta t]\} $
        \State $ D \leftarrow D \cup \{({s}^{s}_{t}, {s}^{l}_{t}, {a}_{t}, r_t, {s}^{s}_{t+1}, {s}^{l}_{t+1})\} $
        \State $ {s}^{s}_{t} \leftarrow {s}^{s}_{t+1}, {s}^{l}_{t} \leftarrow {s}^{l}_{t+1}$
    \EndWhile
    \For{each gradient step}
        \State $ \theta_{i} \leftarrow \theta_{i} - \lambda_{Q} \nabla_{\theta_{i}} J_{Q}(\theta_{i}) $ for $i \in \{1,2\} $
        \State $ \phi \leftarrow \phi - \lambda_{\pi} \nabla_{\phi} J_{\pi}(\phi), \ \alpha \leftarrow \alpha - \lambda_{\alpha} \nabla_{\alpha} J_{\alpha}(\alpha) $
        \State $ \Bar{\theta}_{i} \leftarrow \tau \theta_{i} + (1 - \tau) \Bar{\theta}_{i} $ for $i \in \{1,2\} $
    \EndFor
\EndFor
\Statex $^\dagger$The index of $\mathcal{T}$ is started from 1.
\end{algorithmic}
\end{algorithm}
\setlength{\textfloatsep}{2pt}

\begin{figure*}[t]
\centering
  \includegraphics[width=0.65\textwidth]
  {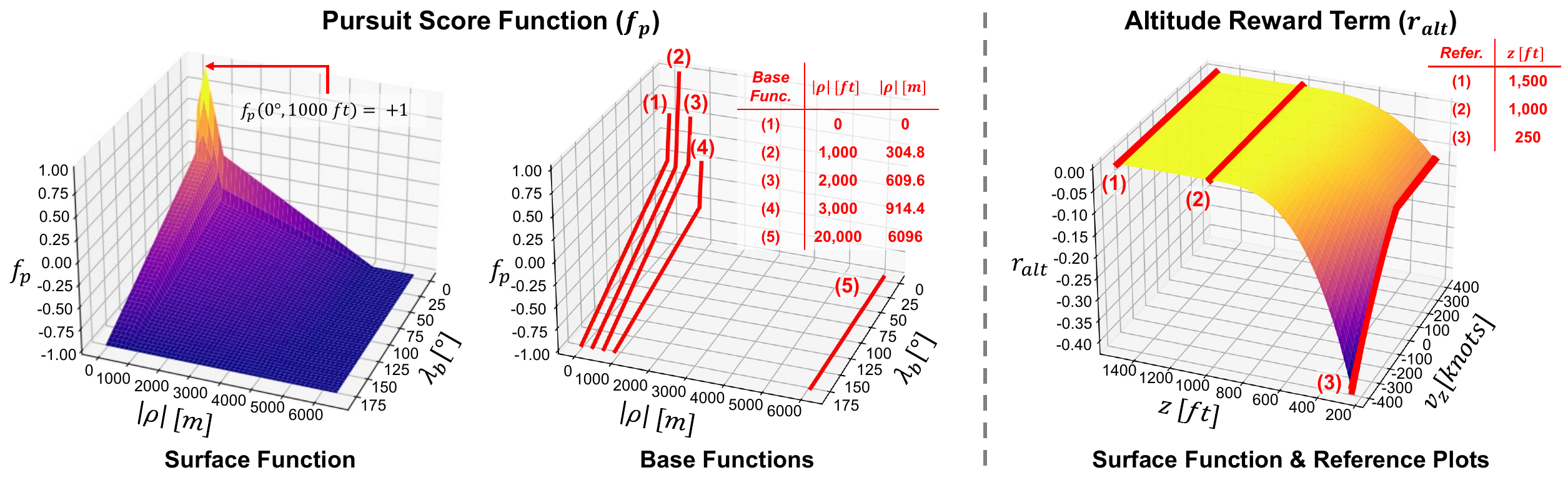}
  \caption{The pursuit score function and the altitude reward term.}
  \label{fig:reward_pursuit}
  \vspace{-0.5em}
\end{figure*}

\begin{table*}[t]
\caption{
Reward terms for aerial dogfights.
}
\makegapedcells
\centering
\begin{adjustbox}{width=0.65\textwidth,pagecenter}
\renewcommand{\arraystretch}{0.2}
\begin{tabular}[t]{ll}
\hline
\multicolumn{1}{c}{Reward Terms (Weight)} & \multicolumn{1}{c}{Reward Functions} \\
\hline
Energy-Pursuit Score (2) &
$
r_{pursuit} = k_{E} \times f_{p}(\lambda_b, \left|\boldsymbol{\rho}\right|), \quad k_{E} =
\begin{cases}
\hfil \left(0.5 + E/E_{max}\footnote{}\right) & \hfil f_{p} \geq 0 \\
\hfil \left(1.5 - E/E_{max}\right) & \hfil f_{p} < 0
\end{cases}
$\\

Horizontal Pursuit (1) &
$
r_{horizon} = - \lambda_e / 180 \degree
$ \\

DZ (Ownship, Opponent) (5) & 
$
r_{DZ-own} = 
\begin{cases}
\hfil +1  & \boldsymbol{\rho} \in \text{DZ}_{own}  \\
\hfil  0  & \boldsymbol{\rho} \notin \text{DZ}_{own}
\end{cases}
$, \quad
$
r_{DZ-oppo} =
\begin{cases}
\hfil -1  & \boldsymbol{\rho} \in \text{DZ}_{oppo}  \\
\hfil  0  & \boldsymbol{\rho} \notin \text{DZ}_{oppo}
\end{cases}
$ \\

Specific Energy (0.5) & 
$
r_{E} = clip((E - E_{des}) / E_{des}\footnote{}, \ -1, \ +1)
$ \\

Altitude (15) & 
$
r_{alt} = \left(k_{alt} \times \text{min}\left(0, \frac{z - z_{low}}{z_{low} - z_{min}}\right) \right)^3, \quad k_{alt} = 0.5 - 0.5 \times \text{min}(0, \frac{{v}_z}{{v}_{max}})
$ \\

AoA (1) & 
$
r_{AoA} =
\begin{cases}
\hfil - \left|\alpha_b\right|/45\degree & \alpha_b < 0\degree \\
\hfil  0  & 0\degree \leq \alpha_b < 45\degree \\
\hfil - (\alpha_b - 45\degree)/45\degree  & \alpha_b \geq 45\degree
\end{cases}
$ \\
\hline
\end{tabular}
\end{adjustbox}
    \begin{tablenotes}
        \footnotesize \centering
        \item[a] $^1E_{max}$ and $^2E_{des}$ are derived from speeds of 800/450 knots and altitudes of 30,000/10,000 feet, respectively, based on \cite{shaw1985fighter}.
    \end{tablenotes}
\vspace{-2.0em}
\label{table:reward}
\end{table*}

Algorithm \ref{algo:sac} summarizes the overall training process.
At each episode, a fixed-length FIFO buffer $\mathcal{T}$ is initialized with $n_l \Delta t$ states observed through actions $a_{init}$ in the environment.
In each step, the state trajectories $s^l_t, s^s_t$ are indexed from the buffer $\mathcal{T}$. The trajectories are then fed to the policy $\pi_{\phi}$ to sample an action $a_t$ for interacting with the environment. After the agent observes the next states $x^{own}_{t+1}, x^{oppo}_{t+1}$ and reward $r_t$, the oldest state is dequeued and $x_{t+1} \! = \{x^{own}_{t+1}, x^{oppo}_{t+1}, a_t\}$ is enqueued in $\mathcal{T}$. New state trajectories are indexed from the updated buffer, and the transition data is stored in a replay buffer $D$.
During the update phase, The model parameters ($\theta_1, \theta_2, \phi$) with the temperature $\alpha$ are updated with the objective functions $J_{Q}(\theta_1)$, $J_{Q}(\theta_2)$, $J_{\pi}(\phi)$ and $J_{\alpha}(\alpha)$. The weight of the target Q-functions $\bar{\theta}_1, \bar{\theta}_2$ are updated by Polyak Averaging\cite{polyak1992acceleration} with a coefficient $\tau$.

\subsection{Reward Function}
The reward function $r_t$ is a weighted summation of the terms described in Table \ref{table:reward}.
Overall, the reward encourages the agent to maximize the energy-pursuit score ($r_{pursuit}$), enhance horizontal pursuit performance ($r_{horizon}$), position the opponent within the ownship's DZ ($r_{DZ-own}$), and avoid the opponent's DZ ($r_{DZ-oppo}$), all while managing the specific energy $E$ close to the desired energy $E_{des}$ ($r_{E}$). Concurrently, it motivates the agent to maneuver within a safe altitude ($r_{alt}$) and the proper AoA ranges ($r_{AoA}$), preventing crashes and aerodynamic stalls.

The energy-pursuit score consists of a pursuit score function $f_{p}$ and a factor $k_{E}$.
The pursuit score function (Fig. \ref{fig:reward_pursuit}) is a surface function constructed by interpolating five piecewise linear base functions. Each base function is defined based on different relative distances ($\left|\boldsymbol{\rho}\right|$) (\SI[group-separator={,},group-minimum-digits=4]{0}-\SI[group-separator={,},group-minimum-digits=4]{20000}{ft}) and includes a breakpoint at a tracking error ($\lambda_b$) of 5 degrees. This two-dimensional function facilitates the agent to reduce the angle with the opponent and to position within the desired relative distance in the damage zone (\SI[group-separator={,},group-minimum-digits=4]{500}-\SI[group-separator={,},group-minimum-digits=4]{3000}{ft}).
Subsequently, the factor $k_{E}$ is determined by the sign of $f_{p}$ and modulates the magnitude of $r_{pursuit}$ by $E$. $k_{E}$ varies between [0.5, 1.0] for a positive $f_{p}$ and [0.5, 1.5] for a negative $f_{p}$, imposing a stricter penalty on the agent under conditions of a negative pursuit score.
This energy-embedded reward term incentivizes the agent to mitigate the loss of specific energy while tracking a target. It encourages the agent to convert altitude and velocity, redistributing potential and kinetic energy, thereby discovering sophisticated maneuvers with robust tracking capability, in line with energy maneuverability theory\cite{shaw1985fighter}.

The altitude term (Fig. \ref{fig:reward_pursuit}) activates from $z_{low}\!=\!\SI[group-separator={,},group-minimum-digits=4]{1000}{ft}$ to prevent the agent from descending below $z_{min}\!=\!\SI{250}{ft}$.
Inspired by the collision term in \cite{seong2021learning}, we design the penalty to become more severe as the size of the negative vertical velocity $v_z$ increases. To amplify this effect, we cube this term, sharply increasing the slope with decreasing altitude and thereby intensifying the change in the penalty. This encourages our agent not only to increase its altitude but also to decrease its descent rate in low altitude situations.

\section{Experiment}
\label{sec:experiment}
\subsection{Environment Setup}
\begin{figure*}[t]
\centering
\includegraphics[width=0.80\textwidth]{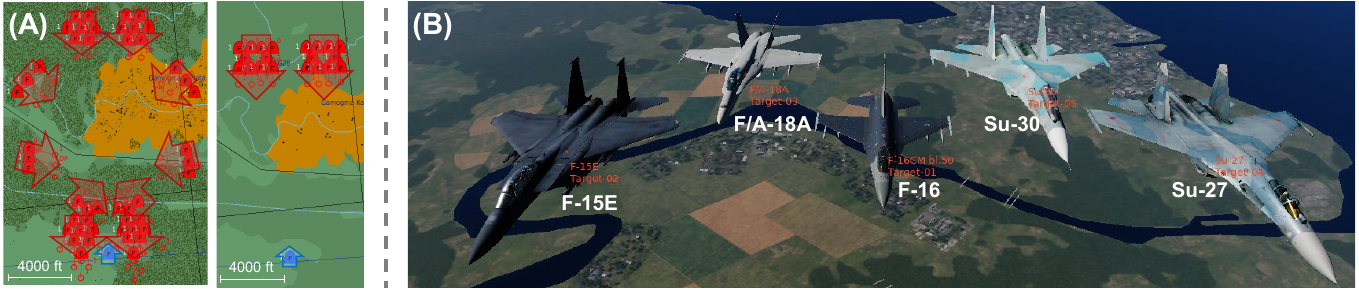}
\caption{
\textbf{(A): }Mission configurations for training (left) and evaluation (right). \textbf{(B): }Different types of aircraft for the opponent.
}
\label{fig:mission_config}
\vspace{-0.5em}
\end{figure*}

We configured dogfight scenarios using DCS that has high-fidelity dynamics and a range of mission configuration tools. We selected the F-16 for our agent, which is an example of a mature fighter jet.
Fig. \ref{fig:mission_config} (A) shows scenario configurations with spawn points and directions.
In training episodes, our agent (blue) was initialized with an altitude of \SI[group-separator={,},group-minimum-digits=4]{15000}{ft} and a speed of \SI{500}{kt}, and the opponent (red), with different altitudes (\SI[group-separator={,},group-minimum-digits=4]{10000}-\SI[group-separator={,},group-minimum-digits=4]{18000}{ft}), was randomly spawned with various locations within \SI[group-separator={,},group-minimum-digits=4]{30000}{ft} from our agent.
In the evaluation, the ownship and opponent were spawned facing each other at a distance of \SI[group-separator={,},group-minimum-digits=4]{10000}{ft}, participating in episodes designed to facilitate fair performance comparisons.
Each episode has the following finishing conditions: 1) if the altitude of any agent falls below $z_{min}$, 2) if the life of any agent drops to 0, 3) if the ownship reaches a maximum of \SI[group-separator={,},group-minimum-digits=4]{9000} time steps, or 4) if agents collide with each other.

We set up various opponent fighters as depicted in Fig. \ref{fig:mission_config} (B). The opponents are operated using a combination of complex BFM-based strategies provided by DCS\cite{DCSalgorithm}. In training, the opponent was randomly selected from four aircraft: one with equivalent specifications (F-16), two with comparable specifications (F-15E, F/A-18A), and one with superior specifications (Su-27) compared to the ownship. Alongside the four aircraft, we further spawned the Su-30 in evaluation, which had not been encountered during training, to assess the robustness of our policy against a new opponent.

\begin{table*}[t] 
\caption{Quantitative Evaluation against 4 Opponents (F-15E, F-16, F/A-18, Su-27) and an Unseen Aircraft (Su-30)}
\makegapedcells
\centering
\begin{adjustbox}{width=0.98\textwidth, pagecenter}
\Large
\renewcommand{\arraystretch}{1.0}
\begin{tabular}{c*{16}{S}|c*{4}{S}}
\toprule
&\multicolumn{4}{c}{{v.s. F-15E}} 
&\multicolumn{4}{c}{{v.s. F-16}}
&\multicolumn{4}{c}{{v.s. F/A-18}}
&\multicolumn{4}{c}{{v.s. Su-27}}
&\multicolumn{4}{|c}{{v.s. Su-30}} \\

\cmidrule(lr){2-5}
\cmidrule(lr){6-9}
\cmidrule(lr){10-13}
\cmidrule(lr){14-17}
\cmidrule(lr){18-21}
Method
    & {\makecell{Win \\ (\%) $\uparrow$}}
        & {\makecell{Loss \\ (\%) $\downarrow$}}
            & {\makecell{Damage \\ (\%) $\uparrow$}}
                & {\makecell{Life \\ (\%) $\uparrow$}}
    & {\makecell{Win \\ (\%) $\uparrow$}}
        & {\makecell{Loss \\ (\%) $\downarrow$}}
            & {\makecell{Damage \\ (\%) $\uparrow$}}
                & {\makecell{Life \\ (\%) $\uparrow$}}
    & {\makecell{Win \\ (\%) $\uparrow$}}
        & {\makecell{Loss \\ (\%) $\downarrow$}}
            & {\makecell{Damage \\ (\%) $\uparrow$}}
                & {\makecell{Life \\ (\%) $\uparrow$}}
    & {\makecell{Win \\ (\%) $\uparrow$}}
        & {\makecell{Loss \\ (\%) $\downarrow$}}
            & {\makecell{Damage \\ (\%) $\uparrow$}}
                & {\makecell{Life \\ (\%) $\uparrow$}}
    & {\makecell{Win \\ (\%) $\uparrow$}}
        & {\makecell{Loss \\ (\%) $\downarrow$}}
            & {\makecell{Damage \\ (\%) $\uparrow$}}
                & {\makecell{Life \\ (\%) $\uparrow$}} \\
\midrule
DCS-Ace   &
\text{34.6} & \text{43.0} & \text{38.9} & \text{54.4} &
\text{38.3} & \text{39.6} & \text{38.8} & \text{59.1} &
\text{18.7} & \text{58.0} & \text{32.9} & \text{32.3} &
\text{22.9} & \text{41.8} & \text{37.0} & \text{52.7} &
\text{ 9.2} & \text{66.4} & \text{19.2} & \text{28.6} \\
MLP   &
\text{37.3} & \text{58.5} & \text{38.2} & \text{40.2} &
\text{ 3.7} & \text{87.6} & \text{ 4.6} & \text{10.1} &
\text{ 6.5} & \text{85.1} & \text{ 8.2} & \text{10.8} &
\text{ 1.7} & \text{31.3} & \text{ 3.8} & \text{53.3} &
\text{ 5.7} & \text{72.4} & \text{ 7.3} & \text{20.0} \\
LSTM   &
\text{88.6} & \text{10.9} & \text{91.8} & \text{89.0} &
\text{28.6} & \text{64.2} & \text{38.5} & \text{34.9} &
\text{83.8} & \text{15.4} & \text{88.5} & \text{79.4} &
\text{55.7} & \text{35.3} & \text{67.2} & \text{60.0} &
\text{60.4} & \text{39.3} & \text{71.3} & \text{58.5} \\
LS-LSTM   &
\text{83.3} & \text{16.7} & \text{86.7}   & \text{78.2} &
\text{75.6} & \text{20.1} & \text{83.2}   & \text{77.5} &
\text{88.1} & \text{11.7} & \textbf{90.8} & \text{87.0} &
\text{82.6} & \text{14.4} & \text{85.9}   & \text{77.0} &
\text{62.4} & \text{36.8} & \text{69.7}   & \text{60.3} \\
TempFuser (w/o LSTM)  &
\text{93.0}   &   \text{ 7.0} &   \text{89.7} & \textbf{93.8} &
\text{83.8}   & \textbf{11.1} &   \text{81.2} & \textbf{93.2} &
\text{88.8}   & \textbf{10.9} &   \text{84.1} & \textbf{90.9} &
\text{82.6}   &   \text{13.9} &   \text{89.0} & \textbf{93.0} &
\text{82.5}   &   \text{14.4} &   \text{82.4} & \textbf{92.3} \\
TempFuser (w/ LSTM, ours)  &
\textbf{93.5} & \textbf{ 6.5} & \textbf{94.2}   & \text{92.5} &
\textbf{86.3} & \text{13.4}   & \textbf{96.0}   & \text{86.6} &
\textbf{89.1} & \textbf{10.9}   & \text{90.3}   & \text{86.2} &
\textbf{92.5} & \textbf{ 7.0} & \textbf{95.1}   & \text{92.0} &
\textbf{86.1} & \textbf{13.9} & \textbf{90.8}   & \text{81.2} \\
\bottomrule

\end{tabular}
\end{adjustbox}
    \label{table:eval}
\vspace{-1.5em}
\end{table*}

\subsection{Baseline Schemes}
\begin{itemize}
    \item 
    \textbf{DCS-Ace: }This is a built-in BFM-based policy with the most challenging skill level, \textit{Ace}, in DCS \cite{DCSalgorithm}.
    
    \item 
    \textbf{MLP: }This is a multilayer perceptron network that observes only the current state $x_t$ (akin to \cite{yoo2022deep}).
    
    \item 
    \textbf{LSTM: }This considers the short-term state trajectory $s^s_t$ only based on an LSTM layer (similar to \cite{bae2023deep}).
    
    \item 
    \textbf{LS-LSTM: }This scheme employs the long short-term temporal input embeddings using $s^l_t, s^s_t$, but consists only of LSTM layers without the transformer encoder.

    \item 
    \textbf{TempFuser (w/o LSTM): }This is an ablation model of our TempFuser, which excludes the LSTM layers but still incorporates the temporal fusion concept.
    
\end{itemize}

\subsection{Evaluation}
\begin{figure*}[t]
\centering
\includegraphics[width=0.95\textwidth]{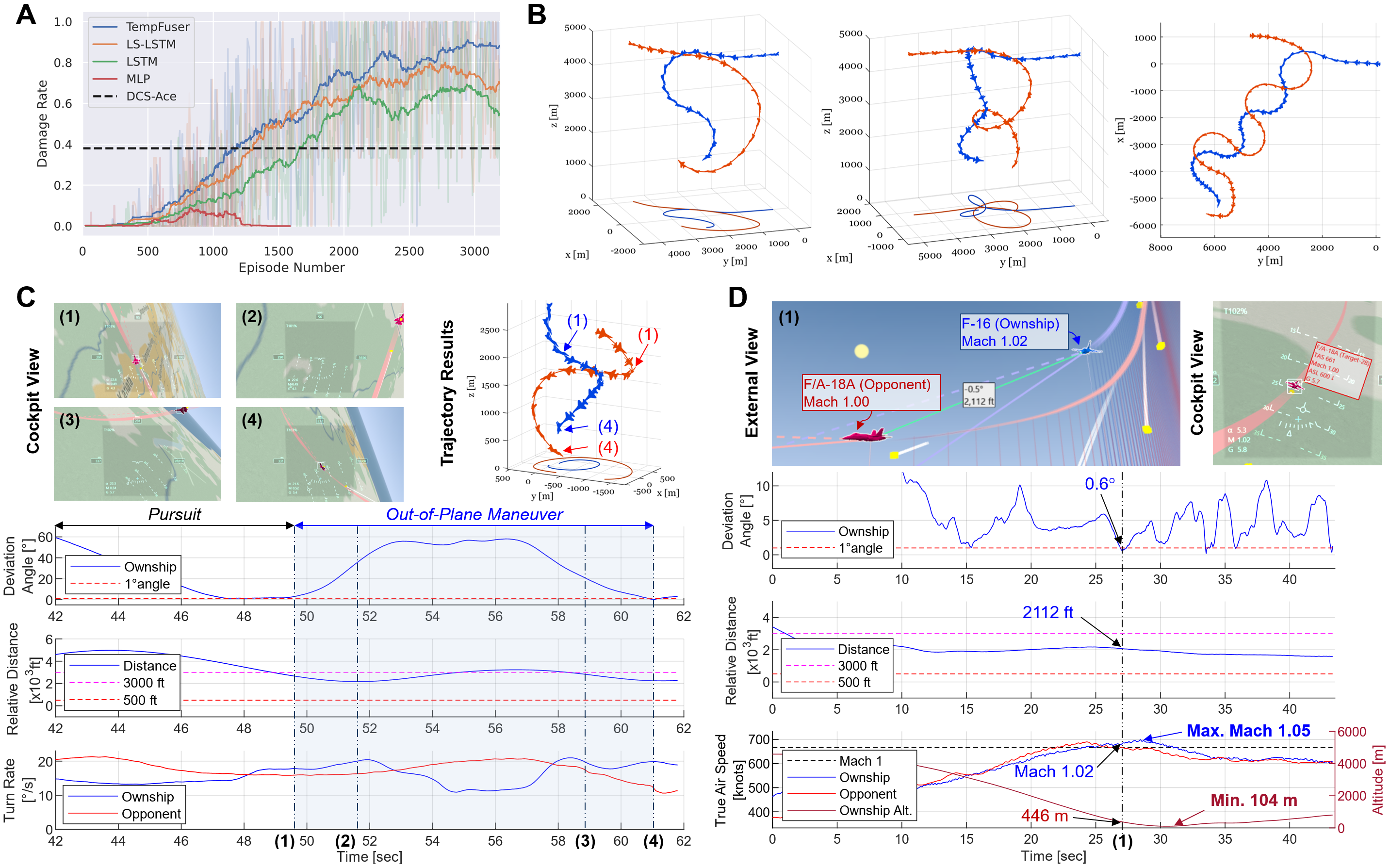}
\caption{
Evaluation results.
\textbf{(A): }Learning curves of the normalized damage rate against the opponent.
\textbf{(B): }Results of engagements against the opponent (left: F-15E, middle: F-16, right: Su-27). 3D flight and horizontally projected trajectories of the ownship (blue) and opponent (red) are illustrated from the beginning until the winning moment.
\textbf{(C): }Quantitative results of learned tactical out-of-plane maneuvers against a superior-spec aircraft (Su-30).
\textbf{(D): }Quantitative results of a near-sonic speed scenario against the F/A-18A opponent. All the cockpit and external views are visualized by Tacview\cite{tacview2023}.
}
\label{fig:eval_result}
\vspace{-1.5em}
\end{figure*}

\textbf{Learning Curves:}
We evaluated learning curves across different policies using the DZ damage ratio. This ratio signifies the proportion of the opponent's life that our agent has reduced through the use of DZ. By assessing this metric once every 10 episodes, we can quantify the improvement in performance over the course of training.
Fig. \ref{fig:eval_result} (A) compares the performance of different baseline policies. The results show that our model surpasses other baselines in terms of performance convergence and learning speed. The MLP was unable to learn the appropriate maneuvers to confront agile opponents.
While the LSTM successfully learned dogfights and achieved a damage rate of over $50\%$—a rate higher than that of the DCS-Ace—its performance plateaued at around $60\%$.
By employing the long and short-term features, the LS-LSTM exhibited better performance compared to the LSTM. However, as it only concatenates these two features without adequately integrating them, it failed to fully leverage their combined potential, preventing it from surpassing the $80\%$ performance threshold.
Incorporating the transformer architecture, TempFuser effectively captured the opponent's tactical and dynamic attributes, allowing an average $90\%$ DZ damage.

\textbf{Quantitative Performance:}
Table \ref{table:eval} summarizes the performance of the different policies against four types of aircraft, as well as an unseen aircraft, the Su-30. We applied four metrics for the quantitative evaluation: win rate (Win), lose rate (Lose), DZ damage rate (Damage), and the life score of the ownship (Life). All the metrics are averaged after each episode.
Our model generally outperforms other baseline models in terms of win and damage rates against different opponent aircraft.
Among the baseline models, the non-learning DCS-Ace often fell short of reaching the end game and struggled to defeat fighters with higher aerodynamic specifications (Su-27/30).
When it comes to the learnable models, the F-16 and Su-27/30 aircraft present the greatest challenge as an opponent. In these scenarios, the ownship is not able to utilize the performance difference between the aircraft based on their aerodynamic specifications.
Only the policy models that incorporate both long and short-term trajectory inputs (LS-LSTM, TempFuser) demonstrate remarkable successes, achieving a win/damage rate of over $50\%$ against the various types of opponents, including those challenging aircraft.
However, the performance of the LS-LSTM was notably diminished when encountering the unobserved opponent, leading to a significant decrease in win/damage rates to around $60\%$.
Our TempFuser method, on the other hand, exhibits only a minor performance decrement while effectively reducing the opponent's life score by an average of $90\%$, achieving the highest win rate ($86\%$) and life rate ($81\%$).
TempFuser (w/o LSTM) also shows superior performance but focuses more on defensive maneuvers (highest life rate) and performs less effectively against rapid opponents (Su-27/30) compared to our model.
This ablation study demonstrates that incorporating local feature extraction of distinct long- and short-term trajectories, followed by global feature fusion using a transformer model, results in better performance than a transformer-only structure.

\subsection{Learned Flight Behavior}
\textbf{Basic and Acrobatic Maneuvers:}
Fig. \ref{fig:eval_result} (B) shows the flight trajectories of our agent against the three different opponents (F-15E, F-16, Su-27) during engagements.
The overall results demonstrate that our method successfully learns to execute complex combinations of basic and acrobatic flight maneuvers\cite{shaw1985fighter}, enabling faster turns and securing an advantage over the opponent.
Fig. \ref{fig:eval_result} (B, left) illustrates the result of an episode against an F-15E opponent. While the opponent performed with an average turn rate of \SI{7.85}\degree/s (max. \SI{16.80}\degree/s), our agent showed a more rapid average turn rate of \SI{11.18}\degree/s (max. \SI{26.41}\degree/s) with a double \textit{'Split-S'} trajectory, enabling it to quickly gain a favorable position and outperform the adversarial agent.

Furthermore, our agent exhibited strategic reactive behavior against the opponent performing complex turns.
Fig. \ref{fig:eval_result} (B, middle) depicts a scenario with the F-16 opponent, which is the same aircraft as our agent. When the opponent performed two turning descents, our model responded by executing a \textit{'Spiral dive'} with a tighter radius and faster turning speed, eventually winning the scenario by targeting the anticipated area where the opponent was expected to reach.
As another scenario, Fig. \ref{fig:eval_result} (B, right) displays the results against an aggressive opponent (Su-27). With superior aerodynamic properties compared to the ownship, the Su-27 performed high-speed flight maneuvers, making pursuit challenging. In response, our agent showed the \textit{'Flat Scissors'} aerobatic maneuver, intersecting the opponent's trajectory and effectively targeting the fast-moving aircraft.

\textbf{Tactical Flight Maneuvers:}
Without explicit prior knowledge, our method discovers and learns tactical maneuvers that are close to the human pilot's skills depicted in Fig. \ref{fig:tactical}.
Fig. \ref{fig:eval_result} (C) illustrates the results of an engaging situation against the Su-30.
Our method executed pursuit maneuvers until 49.7 sec, reducing the tracking error and relative distance by up to \SI{1.4}\degree and \SI[group-separator={,},group-minimum-digits=4]{2621}{ft} (Fig. \ref{fig:eval_result} (C, 1)).
However, the pursuit method alone was not able to place the Su-30, which has a faster average speed of \SI{496}{kt} (\SI{919}{km/h}), in the effective damage zone of our F-16.
To overcome this situation, our policy demonstrated an out-of-plane maneuver, the \textit{`Low Yo-Yo,'} which leveraged gravitational acceleration to perform a rapid turn toward the anticipated path of the opponent, despite the increased tracking angle error (Fig. \ref{fig:eval_result} (C, 1-4)).
These tactical maneuvers enabled our agent to stay within the opponent's turning circle while keeping the distance within \SI{500}-\SI[group-separator={,},group-minimum-digits=4]{3000}{ft}, even against the faster opponent. Moreover, our agent achieved a more rapid instantaneous turn rate (\SI{20.97}\degree/s) than the opponent, strategically placing it within our aircraft's DZ (Fig. \ref{fig:eval_result} (C, 4)).

\textbf{Robust Pursuit in Supersonic Speed: }
We further investigated the robustness of our policy in aggressive scenarios, specifically at supersonic conditions.
Fig. \ref{fig:eval_result} (D) illustrates an aerial scene with overall quantitative results where our agent tracked an F/A-18A opponent evading to a low altitude with near-supersonic velocity.
As the deviation angle decreased, the opponent increased its speed by descending to an altitude below \SI{500}{ft}. Against such a high-speed adversary, our agent maintained the desired distance while traveling at Mach 1.02 (\SI{1259}{km/h}) and reduced the deviation angle by up to 0.6\degree.
It then executed a high-speed pursuit reaching Mach 1.05 (\SI{1297}{km/h}) under critically low altitude conditions (at least \SI{102}{m}), while maintaining a proper distance of around \SI[group-separator={,},group-minimum-digits=4]{2000}{ft} without overshooting.
Finally, it won against the adversary by accumulating damage.
These results show the robustness of our model in tracking agile opponents in supersonic situations while adhering to altitude constraints.
\section{Conclusion}
\label{sec:conclusion}
We introduced TempFuser, a long short-term temporal fusion transformer designed for learning agile and tactical flight maneuvers.
TempFuser integrates long- and short-term temporal transition embeddings to effectively capture the tactics and dynamic features of high-speed aircraft, enabling it to execute acrobatic end-to-end flight controls in agile dogfight scenarios.
In high-fidelity airborne scenarios, the proposed model outperforms other baseline methods, across a diverse range of opponent aircraft types. Our model successfully learns tactical flight maneuvers and robust pursuit strategies without relying on heuristic knowledge.

We believe our work has potential for broader robotic applications beyond dogfights. It could be extended to other agile and interactive scenarios, such as autonomous racing, where understanding the strategies of opponent agents is crucial. In future research, we aim to adapt our method to those fields, as well as to enhance it for multi-agent scenarios beyond the one-versus-one context.
\section*{Acknowledgment}
The authors sincerely thank Prof. Kwangjin Yang for organizing opportunities for us to learn from the valuable hands-on experiences of human pilots at the Korea Air Force Academy.

\ifCLASSOPTIONcaptionsoff
  \newpage
\fi



%
\balance
\bibliographystyle{IEEEtran}
\bibliography{root}

\end{document}